\documentclass[letterpaper, 10 pt, conference]{ieeeconf}  % Comment this line out if you need a4paper

\IEEEoverridecommandlockouts                              % This command is only needed if 
\title{\LARGE \bf
DR-WLC: Dimensionality Reduction cognition for object detection and pose estimation by Watching, Learning and Checking
}

\author{Yu Gao$^{1,2}$, Xi Xu$^{1,2}$, Tianji Jiang$^{1,2}$, Siyuan Chen$^{1,2}$, Yi Yang$^{*,1,2}$, Yufeng Yue$^{1,2}$, Mengyin Fu$^{1,2}$% <-this % stops a space
\thanks{*This work was partly supported by National Natural Science Foundation of China (Grant No. NSFC 61973034, 62233002, U1913203, 61903034 and CJSP
Q2018229)}% <-this % stops a space
\thanks{$^{1}$School of Automation, Beijing Institute of Technology, Beijing, China}%
\thanks{$^{2}$State Key Laboratory of Intelligent Control and Decision of Complex System, Beijing Institute of Technology, Beijing, China}%
\thanks{*Corresponding author: Y. Yang Email: yang yi@bit.edu.cn}%
}

\usepackage{subcaption}
\usepackage{graphicx} %插入图片的宏包
\usepackage{float} %设置图片浮动位置的宏包
\usepackage{cite}
\usepackage{color}
\usepackage{url}
\usepackage{algorithm}
\usepackage{algorithmicx}
\usepackage{stfloats}
\usepackage{amsmath}
\usepackage{threeparttable}
\usepackage{booktabs} % 三线表宏包
\usepackage{array}
\usepackage{multirow} %合并单元格
\usepackage{makecell}

\begin{document}

\maketitle
\thispagestyle{empty}
\pagestyle{empty}

%%%%%%%%%%%%%%%%%%%%%%%%%%%%%%%%%%%%%%%%%%%%%%%%%%%%%%%%%%%%%%%%%%%%%%%%%%%%%%%%
\begin{abstract}

Object detection and pose estimation are difficult tasks in robotics and autonomous driving. Existing object detection and pose estimation methods mostly adopt the same-dimensional data for training. For example, 2D object detection usually requires a large amount of 2D annotation data with high cost. Using high-dimensional information to supervise lower-dimensional tasks is a feasible way to reduce datasets size. In this work, the DR-WLC, a dimensionality reduction cognitive model, which can perform both object detection and pose estimation tasks at the same time is proposed. The model only requires 3D model of objects and unlabeled environment images (with or without objects) to finish the training. In addition, a bounding boxes generation strategy is also proposed to build the relationship between 3D model and 2D object detection task. Experiments show that our method can qualify the work without any manual annotations and it is easy to deploy for practical applications. Source code is at \textcolor[rgb]{0,0,1}{\url{https://github.com/IN2-ViAUn/DR-WLC}}.

\end{abstract}

%%%%%%%%%%%%%%%%%%%%%%%%%%%%%%%%%%%%%%%%%%%%%%%%%%%%%%%%%%%%%%%%%%%%%%%%%%%%%%%%
\section{INTRODUCTION}

Object detection and pose estimation are widely used in unmanned driving, robot navigation and augmented reality. Existing 2D object detection and 3D pose estimation methods based on deep learning often require a large number of 2D image annotation datasets and 3D pose information for deep network training respectively. However, human usually do not have to traverse lots of object images and pose information in various scenes to complete the above tasks. We can reconstruct a 3D model only by simple observations in any scenes to guide us estimate poses and find boundary of objects. Inspired by this idea, we propose a dimensionality reduction cognitive model in this paper. The model can achieve object detection and pose estimation tasks in complex scenes only by 3D model of objects.

The architecture of our model can be divided into three stages: \textbf{Watching, Learning and Checking}. 

In the Watching stage, we first use the designed 3D model observation strategy to generate ground-truth poses information, then EfficientNeRF(E-NeRF) is adopted to generate 2D images of objects by using the ground-truth poses. Finally, the morphological operation is used to obtain bounding boxes. 

In the Learning stage, 2D images from different views are input into the improved Transformer-based object detector DETR-DR to get the prediction poses and bounding boxes. The ground-truth poses and bounding boxes obtained in the Watching stage are used to supervise the training of network. 

\begin{figure*}[htpb]
\centering
\includegraphics[width=\textwidth]{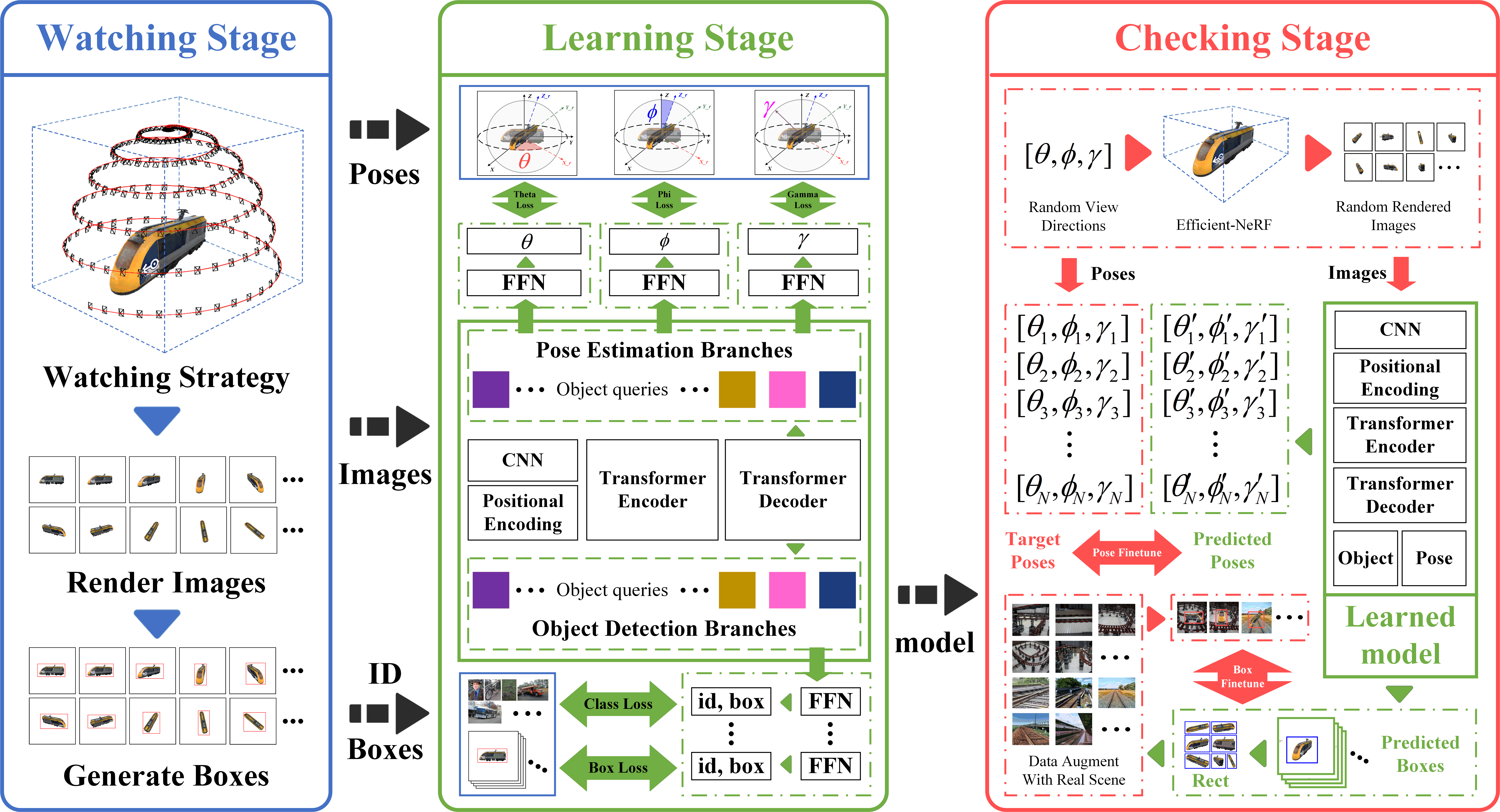}
\caption{\textbf{The pipeline of proposed method.} The architecture of our work consists of three parts. \textbf{The Watching stage} is to generate training data of simple scenes, including raw images with poses and bounding boxes. \textbf{The Learning stage} mainly conducts coarse training of the network to ensure the network has the primary ability to identify objects. In \textbf{the Checking stage}, the background images are used to synthesize complex scenes data. The pre-trained network adopts augment data to finetune weights, which aims to enhance the robustness for complex scenes. It is worth noting that we employ the pre-trained E-NeRF as 3D model repository acting as a supervisor. E-NeRF only participates in forward calculation without any network weight adjustment.}
\label{overview}
\end{figure*}

In the Checking stage, DETR-DR and E-NeRF are combined training to finetune the precision of poses. The unlabeled environment background images are introduced to further improve the robustness of practical environments. Fig. \ref{overview} visualizes this overall pipeline.

Frequent access to 3D model is basic requirement for our method. 3D reconstruction methods based on voxel\cite{voxel-1,voxel-2}, mesh\cite{mesh}, point cloud\cite{point-1,point-2} often take a long time to build the model, and it is difficult to obtain dense and clear rendered images quickly from any views. Significantly, the NeRF\cite{nerf}-series methods greatly improve the quality of rendered results. Although some works focus on improving NeRF towards more accurate\cite{nerf-conv} and larger reconstructed scenes\cite{block-nerf}, the high-resolution rendered results have less effect on object detection that pay more attention to object boundary. The recent works such as EfficientNeRF(E-NeRF)\cite{efficient-nerf}, FastNeRF\cite{fast-nerf} focus on the speed of model, which is a crucial factor to 3D model observation strategy design and implementation. In our method, it is acceptable to give up model rendering accuracy to some extent to greatly improve the rendering speed of images. In addition, since the NeRF-series methods store the model in an implicit way based on neural network, it can be embedded into existing networks, fixing network weights and participating in backpropagation and inference. In this paper, we adopt EfficientNeRF(E-NeRF) as the storage repository for 3D model.

The main contributions of our paper are the following:
\begin{itemize}
\item We propose a dimensionality reduction cognitive model, named DR-WLC, which can perform object detection and pose estimation using only 3D model of objects and unlabeled background images (with or without objects). 

\item A method of generating object ground truth boxes in simple scene is proposed, which build the relationship between 3D model and 2D object detection task.
\end{itemize}

The rest of the paper is organized as follows. Section II provides an overview of object detection methods, NeRF series methods and pose estimation methods. Section III introduces the principles of NeRF series methods and the implementation details of DR-WLC. The experiment results are shown in section IV. Finally, in section V we make a summary of our work.

\section{RELATED WORK}

\subsection{Object Detection Methods}

Object detection is an important research field of computer vision. CNN-based object detectors can be divided into anchor-based detectors and anchor-free detectors. The R-CNN series methods\cite{RCNN, Fast-RCNN, Faster-RCNN, Mask-RCNN}, some of the YOLO series methods\cite{yolov2, yolov3, yolov4, yolov7}, SSD\cite{ssd}, RetinaNet\cite{RetinaNet} all use the anchor mechanism. The newly released YOLOv7\cite{yolov7} surpasses many previous research works in speed and accuracy. However, anchor mechanism not only reduce the generalization ability but also increase the amount of extra computation. Besides, anchor mechanism also lead to problems, such as imbalance of positive and negative samples.

The anchor-free detectors can directly predict objects without pre-setting anchor boxes. Although the researchers pay more attention to anchor-based detectors than anchor-free detectors in the early stage, such as anchor-free mechanism adopted in DenseBox\cite{densebox} and YOLOv1\cite{yolov1} are not inherited in the following works, the design ideas laid the foundation for the future work. In general, anchor-free algorithms can be divided into dense prediction methods\cite{FSAF, FCOS, foveabox, yolox}, and keypoint-based methods\cite{cornernet, centernet, ExtremeNet}. Both of them can achieve the comparable performance as the anchor-based algorithms while avoiding the defects of the anchor mechanism. 

In recent years, with the successful application of Transformer from NLP to CV, some Transformer-based object detectors flourished. DETR\cite{DETR} achieves the successful combination of CNN and Transformer. Deformable DETR\cite{Deformable-DETR} borrowed the idea of deformable convolution and used a multi-scale deformable attention module. These design effectively improved the problems of DETR, such as training speed and poor detection effect on small objects.

\subsection{NeRF Series Methods}

Neural Radiation Field\cite{nerf} adopts neural networks to implicitly describe 3D space. The impressive accuracy and elegant design of NeRF attract great attention. However, NeRF requires a large amount of data and expensive training cost. To improve the speed, it is feasible to change the MLP, which is the most time consuming part. For example, introducing a voxel grid with a small network for implicit storage and rendering\cite{NeRF-2, NeRF-3}, or taking neural 3D point cloud\cite{NeRF-4}, or even abandon the network structure directly\cite{NeRF-5}. Reducing the number of sampling points can also improve rendering speed\cite{efficient-nerf, NeRF-7}. In addition, there are some works that use bake-like methods to achieve acceleration\cite{NeRF-8, NeRF-9,NeRF-10,NeRF-11, NeRF-12}, which transfer the trained model to some structures that can render quickly. Impressive results have also been achieved with the aid of mathematical tools to improve network performance\cite{fast-nerf, NeRF-14}. Reducing the dependence on a large amount of data is also an important research direction for NeRF. There are two common ways, one is to improve the network structure or training method, such as Mip-NeRF\cite{NeRF-15}, infoNeRF\cite{NeRF-16}, etc. The other is to use the scene prior learned by the network, and then perform fine-tuning for the new input\cite{NeRF-17, NeRF-18, NeRF-19, NeRF-20}. Application of NeRF is also an interesting research field. In terms of navigation, NeRF is used as a map, and position estimation is performed based on RGB or depth information\cite{NeRF-21, NeRF-22, NeRF-23, NeRF-24}. At the same time, researchers are also exploring NeRF-related principles and techniques in other fields, such as scene editing\cite{NeRF-25}, style transfer\cite{NeRF-26}.

\subsection{Pose Estimation Methods}
6DoF pose estimation is a key technology for realizing vehicle automatic driving and robot object grasping. The purpose is to complete the transformation from the object coordinate system to the camera coordinate system. The traditional object pose estimation need to use PnP methods to solve the correspondence between 3D-2D point pairs. But since the PnP problem is not differentiable at some points\cite{pose-1}, it becomes difficult to learn all points and weights in an end-to-end manner. Recently, EPro-PnP proposed by Chen et al.\cite{pose-2} makes the PnP problem derivable by introducing a probability density distribution, which greatly enhances the adaptability of the monocular object pose estimation model. The deep learning- based 6DoF pose estimation work no longer relies on 3D template matching\cite{pose-3}. Pose estimation methods based on convolutional neural networks\cite{pose-4, pose-5} and keypoint detection \cite{pose-6, pose-7} have developed rapidly and achieved ideal performance. In addition, pose estimation can also be performed through point cloud information, such as Frustum-PointNet\cite{pose-8} and Voxelnet\cite{pose-9}.

\section{METHOD}

In this section, we first review the basic principles of NeRF and why we choose E-NeRF as the 3D model supervisor. Then, we introduce the details of our method.

\subsection{Background} 
The basic idea of NeRF\cite{nerf} is that given a set of images of a scene with known view directions, a neural network adopts these information for training, and implicitly model the scene. The new view images of the scene are generated by volume rendering. Specifically, NeRF samples 3D point $x$ = ($x$, $y$, $z$) along ray $r$, and takes direction $d$ = ($\theta$, $\phi$) of the sampling point as input. After MLP network (which can be regarded as the mapping function $f$), color $c$ and volume density $\sigma$ corresponding to the sampling point are output. The formula is
$$
(\sigma, c) = f(x, d) \eqno{(1)}
$$

Using the principle of classical volume rendering and the method of numerical approximation, the integral is calculated along each ray passing through the scene. By obtaining the corresponding pixel color on the image plane, a 2D image under the viewing direction is rendered. In addition, NeRF introduces positional encoding, which maps the position and direction information to a high-dimensional space, and then inputs them into the neural network. Positional encoding can effectively solve the poor performance of NeRF in expressing high-resolution complex scenes. In terms of rendering strategy, NeRF adopts "coarse to fine" hierarchical voxel sampling, and samples finer points based on the probability density function generated during coarse stage. NeRF optimizes both coarse and fine networks by minimizing the mean square error of each corresponding pixel color(denoted by $C(r)$ and $\widehat{C}(r)$ respectively) between ground truth image and rendered image, formulated as:
$$
L =  {\sum\limits_{r \in R} {\Vert {C(r) - \widehat{C}(r)} \Vert_2^2} } \eqno{(2)}
$$
where $R$ is the set of all camera rays.

However, NeRF spends a lot of time during training and rendering, which is not conducive to the application. There are some researches currently devoted to improving NeRF in terms of speed\cite{efficient-nerf, NeRF-2, NeRF-3, NeRF-4, NeRF-5,NeRF-7,NeRF-8, NeRF-9,NeRF-10,NeRF-11,NeRF-12}. Among them, E-NeRF\cite{efficient-nerf} improves the training speed through effective sampling and key sampling in coarse stage and fine stage respectively. The designed NerfTree, a 2-depth tree data structure, is employed to cache coarse dense voxels and fine sparse voxels extracted from coarse and fine MLP. This allows for faster inference by directly querying voxels for density and color. E-NeRF can greatly reduce the amount of computation and shorten training and inference time while maintaining accuracy.

\begin{figure}
\centering
\includegraphics[width=0.48\textwidth]{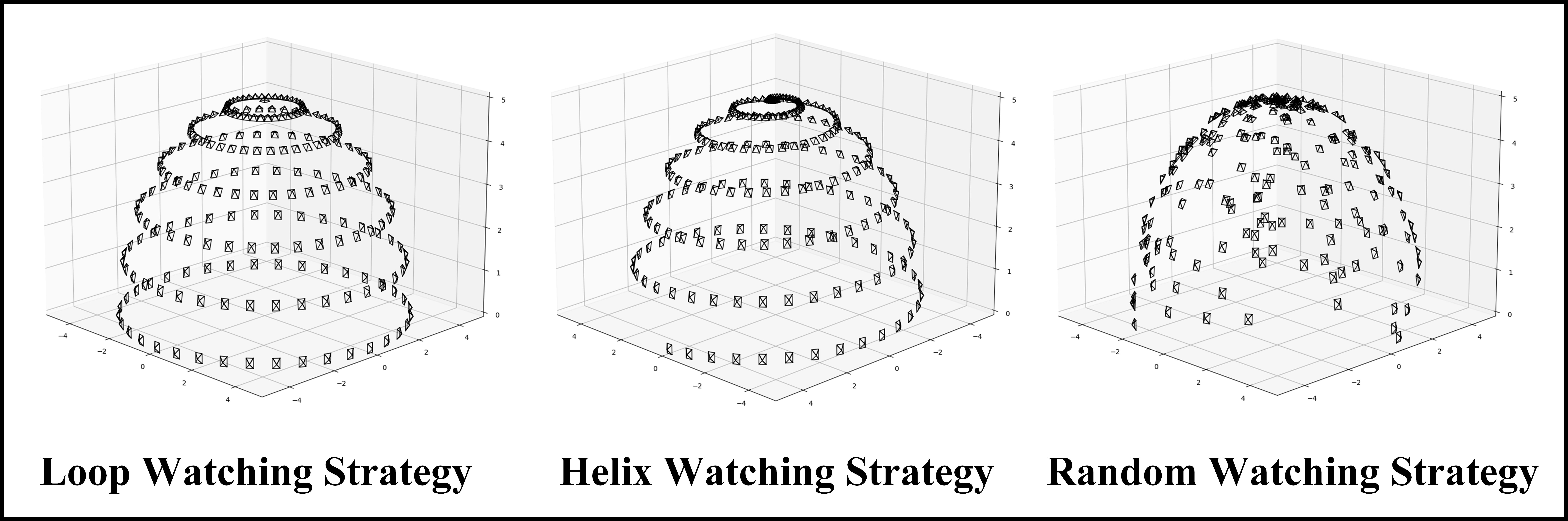}
\caption{The illustration of three watching strategies.}
\label{strategies}
\end{figure}

\begin{figure*}[htpb]
\centering
\includegraphics[width=\textwidth]{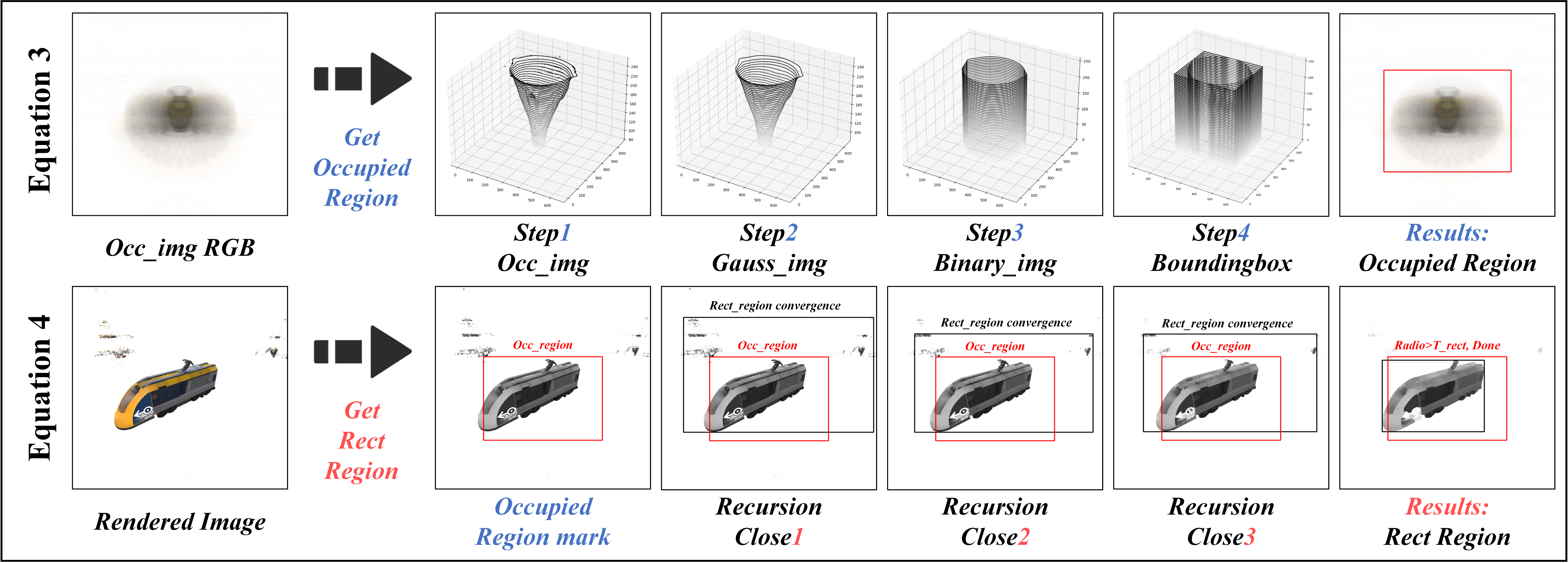}
\caption{\textbf{The generation process of the ground-truth boxes.} The
first line illustrates how to obtain the $occupied \ region$. The second line shows the recursive method for $Rect \ region$.}
\label{bbox}
\end{figure*}

\subsection{Watching Stage} 

There are two main challenges of the Watching Stage:

\begin{itemize}

\item How to design the watching strategy to obtain the ground-truth pose and rendered images. Under the condition of using less data, the performance of DETR-DR is guaranteed.

\item How to automatically generate label of bounding boxes to avoid manual labeling.

\end{itemize}

Fig. \ref{strategies} shows three watching strategies: Loop Strategy, Helix Strategy, and Random Strategy. As defined in NeRF, $\theta$ and $\phi$ represent view direction, while $\gamma$ represents the rendering radius of the E-NeRF model. All watching points are sampled from the 3D sphere formed by $\gamma$. The Loop Strategy divides $\phi$ equally into M parts, where each part corresponds to one watching loop. And $\theta$ is equally divided into N parts. The Helix Strategy divides $\theta$ and $\phi$ equally into N and M watching points respectively. These points are combined in a helix line way to generate a watching route. It should be noted that when evaluating these strategies, it is necessary to ensure these strategies have the same number of watching points. When the watching strategy is determined, pose information ($\theta$, $\phi$ and $\gamma$) and corresponding rendered images generated by E-NeRF are regarded as ground truth labels and enter the Learning stage.

We adopt morphological algorithm to generate label of bounding boxes. Since the background of rendered images obtained by the Nerf series methods is relatively solid, as shown in Fig. \ref{bbox} $Rendered\ Image$, the morphology can meet the requirements of data generation. We superimpose rendered images with weights to obtain occupied region of objects. The process is shown in Equation 3:

$$
\left\{ \begin{array}{l}
Step1:Occ\_img = \sum\limits_{i = 0}^N {({\raise0.7ex\hbox{${img\_RG{B_i}}$} \!\mathord{\left/
 {\vphantom {{img\_RG{B_i}} N}}\right.\kern-\nulldelimiterspace}
\!\lower0.7ex\hbox{$N$}})} \\
Step2:Gauss\_img = Gauss(Gray(Occ\_img))\\
Step3:Binary\_img = Binary(Gauss\_img,T)\\
Step4:Occ\_region = BBox(Binary\_img)
\end{array} \right. \eqno{(3)}
$$

\begin{figure*}[bp]
\centering
$$
\left\{ \begin{array}{l}
{Step1}:Gray\_img = Gray(img\_RGB)\\
{Step2}:Binary\_img = Binary(Close(Gray\_img,kernel),T\_single)\\
{Step3}:temp\_region = BBox(Binary\_img)\\
{Step4}:radio = inter(temp\_region,Occ\_region)/Area(temp\_region)\\
{Step5}:\left\{ \begin{array}{l}
rect\_region = temp\_region,{\rm{ }}radio > T\_rect\\
Gray\_img = Close(Gray\_img),{\rm{ }}kernel +  + {\rm{, }}return\ Step2
\end{array} \right.
\end{array} \right.  \eqno{(4)}
$$
\end{figure*}

Where $N$ represent the images and the total number of images produced by the watching strategy respectively. $Gray$ means converting RGB images to grayscale images. $Gauss$ stands for Gaussian filtering. $Binary$ and $T$ represent binary operation and the threshold. $BBox$ means to calculate the bounding rectangle. $Occ\_region$ is the occupied region. The first line of Fig. \ref{bbox} illustrates the pipeline of algorithm.

The function of the occupied region is to decide whether the boxes generated by each image is available. Although we can obtain solid background with E-NeRF, there is still noise existing due to the E-NeRF performance, as shown in the second line of Fig. \ref{bbox}. The morphological close operation can filter out the interference area, but the noise generated by different images is relatively random, which is difficult to adopt a single threshold for all close operations. Therefore, we use recursive method to increase the size of close kernel gradually to meet the requirements of bounding boxes generation for different images. The process is shown in Equation 4. Where $Close$ represents the morphological close operation. $kernel$ is the size of close operation kernel. $T\_Single$ means the binarization threshold for each image. $inter$ represents the intersection operation, which is the same as it in IoU. $Area$ means the area calculation.

\subsection{Learning Stage}
In the Learning stage, we mainly introduce the DETR-DR model, data augmentation and loss functions. 

DETR-DR is modified by the Transformer-based object detector DETR. As shown in the Learning stage of Fig. \ref{overview}, the classic DETR network only contains object detection branches, which are used to predict classes and bounding boxes. We add the pose estimation branches only by modifying the outputs of DETR for predicting $\theta$, $\phi$, and $\gamma$. An important reason for employing DETR as the basic model is that the complete design stages we proposed have no strict restrictions to the network structure, but have strict restrictions to prediction box format. The object detectors such as YOLO-series methods need NMS to get the final bounding box results, which is difficult to find the corresponding relationship to poses during the training process.

As for data augmentation, we randomly choose the operation of shift, scaling and random background color. In terms of the loss functions, classes prediction and bounding boxes regression adopt cross entropy loss and GIOU loss respectively, which are the same as DETR. And the pose regression is performed by L1 loss.

\subsection{Checking Stage}
The Checking stage focus on two tasks. One is to finetune the weights of pose estimation branches, the other is to enhance the robustness of the object detection branches to the environment. 

The DETR-DR model trained by a small amount of ground-truth pose generated in the Watching stage have poor performance for practical applications. In the Checking stage, E-NeRF and DETR-DR models are jointly trained. Firstly, a large amount of random poses feed into E-NeRF to get rendered images. Secondly, the images are fed into  DETR-DR to predict poses and bounding boxes. Finally, the pose estimation performance is finely optimized by the loss between the predicted pose and the random pose. In the pose finetune loss function, $\gamma$ still uses L1 loss. $\theta$ and $\phi$ are regarded as the width and height of image respectively, and the IoU loss is calculated as shown in Equation 5.

$$
\left\{ \begin{array}{l}
Inter = \min ({\theta _{prd}},{\theta _{trt}}) \cdot \min ({\phi _{prd}},{\phi _{trt}})\\[3mm]
L{\rm{oss}} = 1 - \displaystyle{\frac{{Inter}}{{{\theta _{prd}} \cdot {\phi _{prd}} + {\theta _{trt}} \cdot {\phi _{trt}} - Inter}}}
\end{array} \right.  \eqno{(5)}
$$

where ${\theta_{prd}}$ and ${\phi_{prd}}$ represent the predicted poses. ${\theta_{trt}}$ and ${\phi_{trt}}$ represent the ground-truth poses.

In order to enhance the robustness of complex scenes, a large number of scene images are introduced as background without any annotation. The bounding boxes surrounded regions predicted by DETR-DR are cut out to generate $Rect$. The $Rect$ are randomly shifted, scaled, and fused with background images to generate a new dataset. Since the regions unoccupied by objects of $Rect$ are solid, the fusion can be done with morphological and AND operation, which only has small impact on the scene images. It should be noted that the above composite images are only used for training the object detection branches and do not participate in pose finetune. The object detection branches fine-tuning loss function is GIOU.

\begin{figure}
\centering
\includegraphics[width=0.48\textwidth]{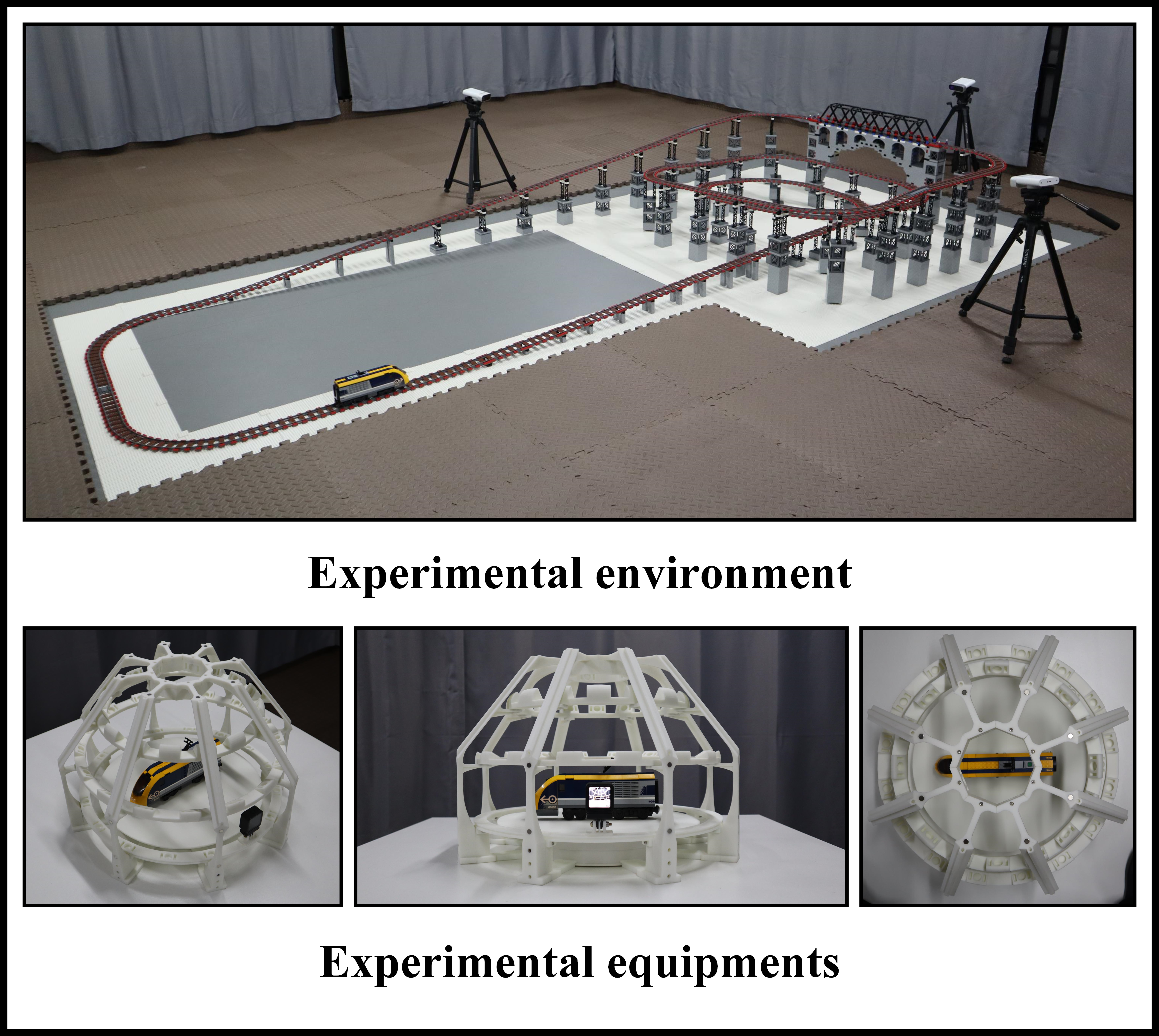}
\caption{Experimental environment and equipments.}
\label{environment and equipments}
\end{figure}

\section{EXPERIMENTS AND RESULTS}

In this section, we explain the details in the experiments and present the results. 

\subsection{Experimental Environment and Equipments}

We adopt the locomotive in LEGO-60197 as the experimental model. The designed special data acquisition equipment is used to collect the training data of E-NeRF. The equipment consists of three observation loops. The interval angle $\phi$ of each observation loop is 30°, and there are 16 camera view points in each loop with an interval $\theta$ of 22.5°. The top layer of the equipment is a regular octagonal structure with 4 camera view points at 90° intervals. We designed and built the LEGO rail viaduct as experimental environment. Data are acquired using the RGB mode of Kinect Azure cameras. Specifically as shown in Fig. \ref{environment and equipments}.

\subsection{Watching Strategy Experiment}

We analyze the effects of three watching strategies. The evaluation results are shown in Table \ref{tab:1}. \textbf{View Points} represents the number of sampled viewpoints. \textbf{Step} means the resolution of angle. \textbf{mAP} (mean Average Precision) is used to evaluate object detection performance. \textbf{Ave deg} represents the average angle error to evaluate the results of pose regression. $\gamma$ is fixed and equal to the observation radius of E-Nerf.

We use 1800 images randomly generated by E-Nerf as evaluation dataset. The method employed to generate bounding boxes is the same as it in the Watching stage. The specific parameters: $T\_occ = 0.95 \cdot max(Gauss\_img) = 242.25$, $T\_single = 0.9 \cdot max(Gray\_img) = 229.5$,  $T\_rect = 0.75$. The training epochs of all strategies are 1000 with the same learning rate of 0.001. 

Four groups of comparative experiments with different parameters are adopted to evaluate the Watching strategy. With the increase of viewpoints number, the performance of three strategies are all improved, which means the number of viewpoints is a decisive factor. We use global average method to comprehensively evaluate these strategies under the condition of the same number of viewpoints. As shown in the blue font of Table \ref{tab:1}, the Random strategy has great advantage.

\begin{table}[htbp]
    \centering
    \caption{The evaluation results of three watching strategies.}
    \label{tab:1} 
    \tabcolsep=0.15cm  % 列间距
    \renewcommand{\arraystretch}{1.5}  % 行高
    \begin{tabular}{ccccccc} 
    \hline\hline\noalign{\smallskip}
    \multirow{2.5}{*}{\textbf{Strategy}} & \multirow{2.5}{*}{\textbf{View Points}} & \multirow{2.5}{*}{\makecell[c]{\textbf{Step (deg°)} \\  $\theta$ $\in [0, 360)$ \\  $\phi$ $\in [0, 90)$ } } & \multirow{2.5}{*}{\textbf{mAP}}&\multicolumn{3}{c}{\textbf{Ave deg (deg°)}}\\
    \cmidrule{5-7}
    & & & &$\theta$ &$\phi$ &$\gamma$ \\
    \midrule
    Loop   & 144 & \multirow{3}{*}{\makecell[c]{$\theta$ step: 10 \\ $\phi$ step: 30}}  & 0.558  & 37.445   & 6.177  & 0.694  \\
    Helix  & 144 &                                                                      & 0.577  & 35.100   & 6.911  & 0.705  \\
    Random & 144 &                                                                      & 0.509  & 23.961   & 3.460  & 0.541  \\

    \midrule
    Loop   & 288 &  \multirow{3}{*}{\makecell[c]{$\theta$ step: 5 \\ $\phi$ step: 30}}  & 0.523  & 17.235   & 5.881  & 0.610  \\
    Helix  & 288 &                                                                      & 0.576  & 20.037   & 5.459  & 0.652  \\
    Random & 288 &                                                                      & 0.584  & 15.961   & 2.296  & 0.568  \\
    
    \midrule
    Loop   & 252 &  \multirow{3}{*}{\makecell[c]{$\theta$ step: 10 \\ $\phi$ step: 15}}  & 0.517  & 15.475   & 3.450  & 0.628  \\
    Helix  & 252 &                                                                       & 0.432  & 19.286   & 2.203  & 0.652  \\
    Random & 252 &                                                                       & 0.586  & 21.657   & 2.465  & 0.542  \\
    
    \midrule
    Loop   & 504 &  \multirow{3}{*}{\makecell[c]{$\theta$ step: 5 \\ $\phi$ step: 15}}  & 0.472  & 11.806   & 4.208  & 0.602  \\
    Helix  & 504 &                                                                      & 0.650  & 6.981   & 3.163  & 0.615  \\
    Random & 504 &                                                                      & 0.523  & 10.371   & 2.327  & 0.583  \\
    \end{tabular}
    
    \tabcolsep=0.14cm
    \renewcommand{\arraystretch}{1.5}
    \begin{tabular}{ccccc} 
    \hline\hline\noalign{\smallskip}
    \multirow{1}{*}{\textbf{Strategy}} & \multirow{1}{*}{\textbf{Global mAP}} & \multirow{1}{*}{\textbf{Global Ave-$\theta$}} & \multirow{1}{*}{\textbf{Global Ave-$\phi$}}&\multirow{1}{*}{\textbf{Global Ave-$\gamma$}}\\
   
    \midrule
    Loop   & 0.518                        
           & 20.490
           & 4.929        
           & 0.634    \\
    Helix  & \textcolor[rgb]{0,0,1}{\textbf{0.559}} 
           & 20.351                                 
           & 4.434        
           & 0.656    \\
    Random & 0.551                           
           & \textcolor[rgb]{0,0,1}{\textbf{17.988}}       
           & \textcolor[rgb]{0,0,1}{\textbf{2.637}}        
           & \textcolor[rgb]{0,0,1}{\textbf{0.559}}    \\
    
    \bottomrule
    \end{tabular}
\end{table}

\begin{table}[htbp]
    \centering
    \caption{The evaluation results of different stages.}
    \label{tab:2} 
    \tabcolsep=0.08cm  % 列间距
    \renewcommand{\arraystretch}{1.5}  % 行高
    \begin{tabular}{cccccc} 
    \hline\hline\noalign{\smallskip}
    \multirow{2.5}{*}{\textbf{After Stage}} & \multirow{2.5}{*}{\textbf{Strategy}} & \multirow{2.5}{*}{\textbf{mAP}}&\multicolumn{3}{c}{\textbf{Ave deg (deg°)}}\\
    \cmidrule{4-6}
    & & &$\theta$ &$\phi$ &$\gamma$ \\
    \midrule
    \multirow{3}{*}{Learning}  & Loop[$\theta$:2°, $\phi$: 5°]  & 0.772  & 9.676   & 3.222  & 0.582  \\
                               & Helix[$\theta$:2°, $\phi$: 5°] & 0.831  & 9.789   & 2.017  & 0.518  \\
                               & Random][All: 3420]             & 0.804  & 7.018   & 1.896  & 0.613  \\

    \midrule
    \multirow{3}{*}{Checking}  & Loop][$\theta$:2°, $\phi$: 5°]+R[5000]     & 0.798  & 5.959   & 1.864  & 0.615  \\
                               & Helix][$\theta$:2°, $\phi$: 5°]+R[5000]    & 0.843  & 3.261   & 1.156  & 0.606  \\
                               & Random][$\theta$:2°, $\phi$: 5°]+R[5000]   & \textcolor[rgb]{0,0,1}{\textbf{0.850}}  & \textcolor[rgb]{0,0,1}{\textbf{1.981}}   & \textcolor[rgb]{0,0,1}{\textbf{0.977}}  & \textcolor[rgb]{0,0,1}{\textbf{0.507}}  \\
    \bottomrule
    \end{tabular}
    
\end{table}

\subsection{Detection and Pose Estimation Evaluation}

In order to verify the fine-tuning effect of the Checking stage, we compare the performance of DETR-DR pass through different stages, as shown in Table \ref{tab:2}. \textbf{Strategy} represents the combination of strategies adopted by DETR-DR. \textbf{R} means the number of random view points chose during the joint training process. $\gamma$ is fixed and equal to the observation radius of E-Nerf.

We generate 5400 new images randomly by E-NeRf as evaluation dataset. The dataset is different from it used in Watching Strategy Experiment, but the parameters in data generation and training process are the same as those in Watching Strategy Experiment.

Compared with Table \ref{tab:1}, the trend of indicators in Table \ref{tab:2} is relatively clear.  the performance of random strategy is still the best with the re-training process provided by Checking stage, both the performance of object detection and pose estimation are improved greatly. The results show that the Checking stage is effective and necessary.

\subsection{Detection and Pose Estimation in real scene}

\begin{figure}
\centering
\includegraphics[width=0.48\textwidth]{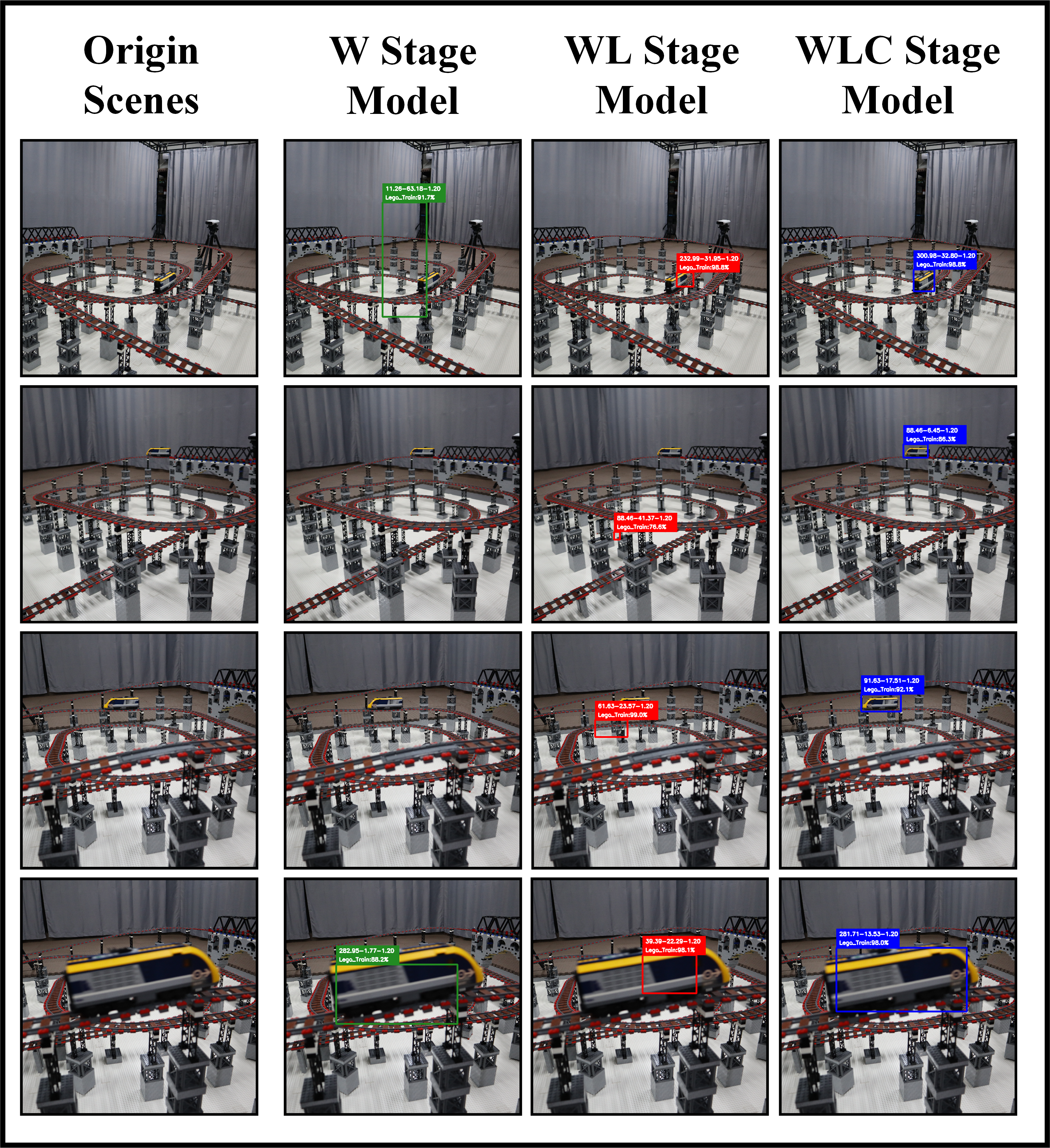}
\caption{\textbf{Detection results in real scenes generated by DETR-DR trained on different stages.} The first column shows original images. The second column shows the results provided by the model only trained with E-NeRF rendered images without data augmentation (the same training strategy with Table \ref{tab:1} experiment). The third and last columns show the results after the Learning and checking stage respectively.}
\label{results}
\end{figure}

Different from the object detection method trained on public datasets (COCO, KITTI, PASCAL VOC), our method is not able to obtain all 3D models of detected objects in the datasets. This makes it difficult to use the public datasets for evaluation. In addition, our training process have no human-labeled annotations, which means there are no ground-truth poses and boxes of objects in the test scene. Therefore, we only present the prediction results as shown in Fig. \ref{results}.

\section{CONCLUSION}

In this work, we propose a dimensionality reduction cognitive model, named DR-WLC, which can perform object detection and pose estimation at the same time. The method only adopt 3D model of objects and unlabeled environment images (with or without objects) as the datasets for training, and make it suitable for practical applications. Inspired by the implicit network structure of NeRF-series models, we successfully embed EfficientNeRF as a supervisor into the complete training process to achieve joint training. In addition, we propose a method to generate ground-truth boxes for NeRF-series rendering images. The basic relationship between 3D model and 2D object detection task is built. The experiments demonstrate the effectiveness of DR-WLC, and also prove that our method is feasible in engineering.

% \addtolength{\textheight}{-12cm}   

%%%%%%%%%%%%%%%%%%%%%%%%%%%%%%%%%%%%%%%%%%%%%%%%%%%%%%%%%%%%%%%%%%%%%%%%%%%%%%%%

%%%%%%%%%%%%%%%%%%%%%%%%%%%%%%%%%%%%%%%%%%%%%%%%%%%%%%%%%%%%%%%%%%%%%%%%%%%%%%%%



\section*{ACKNOWLEDGMENT}

The authors would like to thank Jiadong Tang, Zhaoxiang Liang, Hao Liang, Daiwei Li and all other members of ININ Lab of Beijing Institute of Technology for their contribution to this work.

%%%%%%%%%%%%%%%%%%%%%%%%%%%%%%%%%%%%%%%%%%%%%%%%%%%%%%%%%%%%%%%%%%%%%%%%%%%%%%%%

\bibliographystyle{ieeetr}
\bibliography{reference}

\end{document}